\documentclass[sigconf]{acmart-me}

\usepackage{booktabs} 
\usepackage{url}
\usepackage{color}
\usepackage{enumitem}
\usepackage{balance}

\setcopyright{rightsretained}

\acmDOI{}

\acmISBN{}

\acmConference[MediaEval'21]{Multimedia Evaluation Workshop}{December 6-8 2021}{Bergen, Norway and Online} 
\acmYear{2021}
\copyrightyear{}

\acmPrice{}

\begin{document}
\title{Visual Sentiment Analysis: A Natural Disaster\\ Use-case Task at MediaEval 2021}

\author{Syed Zohaib Hassan\textsuperscript{1}, Kashif Ahmad\textsuperscript{2}, Michael A. Riegler\textsuperscript{1}, \\ Steven Hicks\textsuperscript{1}, Nicola Conci\textsuperscript{3}, Pål Halvorsen\textsuperscript{1}, Ala Al-Fuqaha\textsuperscript{2}}
\affiliation{\textsuperscript{1}SimulaMet, Norway \\ \textsuperscript{2}Information and Computing Technology (ICT) Division, College of Science and Engineering, Hamad Bin Khalifa University, Doha 34110, Qatar \\ \textsuperscript{3}University of Trento, Italy}
\email{ syed@simula.no, kahmad@hbku.edu.qa, michael@simula.no, steven@simula.no, }
\email{ nicola.conci@unitn.it, paalh@simula.no,  aalfuqaha@hbku.edu.qa}

\renewcommand{\shortauthors}{Hassen et al.}
\renewcommand{\shorttitle}{Visual Sentiment Analysis}

\begin{abstract}
The \textit{Visual Sentiment Analysis} task is being offered for the first time at MediaEval. The main purpose of the task is to predict the emotional response to images of natural disasters shared on social media. Disaster-related images are generally complex and often evoke an emotional response, making them an ideal use case of visual sentiment analysis.  We believe being able to perform meaningful analysis of natural disaster-related data could be of great societal importance, and a joint effort in this regard can open several interesting directions for future research. The task is composed of three sub-tasks, each aiming to explore a different aspect of the challenge. In this paper, we provide a detailed overview of the task, the general motivation of the task, and an overview of the dataset and the metrics to be used for the evaluation of the proposed solutions.

\end{abstract}

\maketitle

\section{Introduction}\label{sec:intro}
An enormous amount of multimedia content is generated and shared over the internet daily, especially after the introduction of several social media platforms, such as Twitter, Instagram, and Facebook.
Among other types of content, visual content is used extensively to deliver a specific message to the viewer, implying the common phrase “a picture is worth a thousand words” \cite{said2019natural}. Various emotions can be expressed in visual content, where extracting and analyzing them could be of great value in different application domains, such as education, entertainment, advertisement, and journalism. However, it is not clear which part of an image evoke certain emotions and, more importantly, how the underlying sentiments can be derived from a scene by an automatic algorithm and how the sentiments can be expressed. This opens an interesting line of research to interpret emotions and sentiments perceived by users viewing visual content.

Disaster-related images can be challenging to interpret, especially when analyzing them using current visual analysis techniques. These images generally contain important details in both the background and foreground, making no single part the main point of interest \cite{khan2021explainable}. Developing a visual sentimental analysis pipeline for such a use-case is difficult but could greatly serve our society. Example use cases include having news agencies fact check and provide their audience with relevant information in case of adverse events. It can also be used to assess the damage caused by a disaster and use the relevant imagery to create awareness and raise funds for humanitarian activities. We note that the resultant solutions are not intended to manipulate public emotions rather are concerned with estimating disaster severity and supporting disaster relief. For instance, humanitarian organizations can benefit from these solutions to reach a wider audience by communicating visual content that best demonstrates the evidence of a certain event \cite{Zohaib2020}.

In order to facilitate future work in the domain, a large-scale dataset is collected, annotated, and made publicly available. For the annotation of the dataset, a crowd-sourcing activity with a large number of participants has been conducted.

\section{Related Work}\label{sec:work}
The literature reports several interesting works aiming to minimize and mitigate the loss of lives and infrastructure damage. Different types of solutions rely on ground sensors and satellite imagery for effective disaster handling~\cite{Adeel2018}. With the emergence of social media, more human-centric approaches have arisen for effective response management to natural disasters. There are also some efforts on sentiment analysis of disaster-related text. For instance, IN-SPIRE~\cite{Gregory2007} and PUlSE~\cite{Smith2001} are two frameworks dealing with the sentimental analysis of text data shared on social media.There has been a great deal of work done on textual sentimental analysis of disaster-related textual content. However, the concept of extracting sentiments from the visual content is quite new.  There are also some efforts on visual sentiment analysis of natural disaster-related multimedia content. For instance, in \cite{Zohaib2019,Zohaib2020} a deep visual sentiment analysis is introduced along with a benchmark dataset visual sentiment analysis of natural disaster analysis. We believe the task will help in developing a task force to further explore this interesting aspect of natural disaster-related visual content. 

\section{Dataset Details}
As a first step, in creating this benchmark dataset~\cite{Zohaib2020}, we crawled images related to natural disasters from the social media platforms, such as Twitter, Google API, and Flicker. During this process, special consideration was put on the fact that the crawled images meet the licensing policy of free usage and sharing. A keyword search was the basic technique used to crawl these images, for example, searching by type of disaster, such as \textit{floods}, \textit{wildfires}, \textit{earthquake}, \textit{landslides}, and \textit{storms}, or some specific recent natural disasters that happened in different parts of the world, such as \textit{floods in Pakistan} etc.

After data collection, the next step was data annotation, which was crowd-sourced using Microworkers. The challenging part for crowd-sourcing was designing the task and selecting proper keywords to represent different categories of the annotated data. The three keywords namely \textit{Positive}, \textit{Negative}, and \textit{Neutral} are the most widely used sentiment labels in the literature. However considering the applications and the complexity of sentimental analysis of disaster-related images, it is quite complicated to interpret and represent the sentiment and emotion in these three categories. We rather need a more specific and larger set of sentiment categories/labels to interpret the emotions associated with natural disasters. There is an interesting recent work in psychology on emotions and sentiments representation, where 27 different emotion categories are reported. Based on this study, we set up four different types of annotations, which are expected to better represent the sentiments and emotions associated with natural disasters by covering different aspects of natural disasters. These sets of labels are described below.

\begin{enumerate}
  \item Positive, negative, and neutral.
  \item Relax/calm, normal and simulated/excited.
  \item Joy, sadness, fear, disgust, anger, surprise, and neutral.
  \item Anger, anxiety, craving, empathetic pain, fear, horror, joy, relief, sadness, and surprise.
\end{enumerate}
As a first step towards visual sentiment analysis of natural disaster multimedia content, the scope of the task is limited to images from different types of natural disasters. Overall, $4,003$ images were selected for the crowd-sourcing study. To ensure quality and consistency, each image was annotated by five different persons. The study was disseminated through multiple channels. A total of $10,010$ responses were received during the study, and $2,338$ participants participated from $98$ different countries.

\section{Task Description}\label{sec:approach}
It is the first time we are hosting this task in MediaEval. The task is closely related to previous tasks namely ''The 2017-2019 Multimedia Satellite Task:
Emergency Response for Flooding Events'' \cite{bischke2017multimedia} and ''The 2020 Flood-related Multimedia Task'' \cite{andreadis2020flood}. However, the goals, challenges, images, and the types of natural disasters covered in the dataset are different from the previous tasks.

Images from the dataset provided for the task are filtered to check whether there is a majority consensus among the annotations provided by the crowd-sourcing participants. The resulted dataset consists of $3,631$ images. The development set is composed of $2,432$ images, and the test set contains $1,199$ images. The task is divided into three subtasks, which are described below.

\subsection{Subtask 1}
This is a multi-class single label classification task, where the images are arranged in three different classes, namely \textit{positive}, \textit{negative}, and \textit{neutral}. There is a strong imbalance towards the negative class, given the nature of the topic.

\subsection{Subtask 2}
This is a multi-class multi-label image classification task, where the participants are provided with multi-labeled images. The multi-label classification strategy, which assigns multiple labels to an image, better suits our visual sentiment classification problem and is intended to show the correlation of different sentiments. In this task, seven classes, namely \textit{joy}, \textit{sadness}, \textit{fear}, \textit{disgust}, \textit{anger}, \textit{surprise}, and \textit{neutral}, are covered.

\subsection{Subtask 3}
This task is also multi-class multi-label, however, a wider range of sentiment classes are covered. Going deeper in the sentiment hierarchy, the complexity of the task increases. The sentiment categories covered in this task include \textit{anger}, \textit{anxiety}, \textit{craving}, \textit{empathetic pain}, \textit{fear}, \textit{horror}, \textit{joy}, \textit{relief}, \textit{sadness}, and \textit{surprise}.

\section{Evaluation}
The proposed solutions will be evaluated using a weighted $F_1$ score. It is a more preferable metric when dealing with imbalanced datasets as it is more sensitive to data distribution. It also provides a good balance between precision and recall as it is a harmonic mean of these two metrics. The weighted $F_1$ score can be calculated using the following four equations.
\begin{equation}\label{eq:prec}
Precision_{i} = \frac{\text{True Positives}_i}{(\text{True Positives}_i+\text{False Positives}_i)}
\end{equation}
\begin{equation}\label{eq:rec}
Recall_{i} = \frac{\text{True Positives}_i}{(\text{True Positives}_i+\text{False Negatives}_i)}
\end{equation}
\begin{equation}\label{eq:f1}
\text{$F_1$ Score}_{i} = \frac{2*Precision_{i} *Recall_{i} }{(Precision_{i} +Recall_{i} )}
\end{equation}
\begin{equation}\label{eq:wf1}
Weighted\ F_1 = \frac{\sum_{i=1}(\text{$F_1$ Score}_{i} *c_i)}{\sum_{i=1}(c_i)}
\end{equation}

Equation 1, 2, and 3 shows the calculation of precision, recall and $F_1$ score for each class, respectively. Equation 4 shows the calculation weighted $F_1$ score where $c_i$ is the number of instances of a particular class.

\section{Discussion and Outlook}
Though the literature reports some initial efforts on the topic, several important aspects need to be addressed yet. We believe that extracting and representing sentiments from disaster-related visual content will benefit several stakeholders, such as news agencies, Government and non-government, and other humanitarian organizations. From a research point of view, visual sentiment analysis of natural disaster-related visual content is not limited to object recognition. Instead, it requires a more complex identification analysis and establishing a connection among salient objects, scenes, expressions, and color schemes. We hope the task will result in a joint effort towards this important topic, and it also will open new research directions.

\bibliographystyle{ACM-Reference-Format}
\def\bibfont{\small}
\bibliography{sigproc.bib} 


\begin{thebibliography}{00}


\ifx \showCODEN    \undefined \def \showCODEN     #1{\unskip}     \fi
\ifx \showDOI      \undefined \def \showDOI       #1{#1}\fi
\ifx \showISBNx    \undefined \def \showISBNx     #1{\unskip}     \fi
\ifx \showISBNxiii \undefined \def \showISBNxiii  #1{\unskip}     \fi
\ifx \showISSN     \undefined \def \showISSN      #1{\unskip}     \fi
\ifx \showLCCN     \undefined \def \showLCCN      #1{\unskip}     \fi
\ifx \shownote     \undefined \def \shownote      #1{#1}          \fi
\ifx \showarticletitle \undefined \def \showarticletitle #1{#1}   \fi
\ifx \showURL      \undefined \def \showURL       {\relax}        \fi
\providecommand\bibfield[2]{#2}
\providecommand\bibinfo[2]{#2}
\providecommand\natexlab[1]{#1}
\providecommand\showeprint[2][]{arXiv:#2}

\bibitem[\protect\citeauthoryear{Adeel, Gogate, Farooq, Ieracitano, Dashtipour,
  Larijani, and Hussain}{Adeel et~al\mbox{.}}{2018}]%
        {Adeel2018}
\bibfield{author}{\bibinfo{person}{Ahsan Adeel}, \bibinfo{person}{Mandar
  Gogate}, \bibinfo{person}{Saadullah Farooq}, \bibinfo{person}{Cosimo
  Ieracitano}, \bibinfo{person}{Kia Dashtipour}, \bibinfo{person}{Hadi
  Larijani}, {and} \bibinfo{person}{Amir Hussain}.}
  \bibinfo{year}{2018}\natexlab{}.
\newblock \showarticletitle{A NewEfficient Alert Model for Disaster
  Management}.
\newblock \bibinfo{journal}{{\em Geological Disaster Monitoring Based on Sensor
  Networks\/}} (\bibinfo{year}{2018}), \bibinfo{pages}{57–66}.
\newblock


\bibitem[\protect\citeauthoryear{Andreadis, Gialampoukidis, Karakostas,
  Vrochidis, Kompatsiaris, Fiorin, Norbiato, and Ferri}{Andreadis
  et~al\mbox{.}}{2020}]%
        {andreadis2020flood}
\bibfield{author}{\bibinfo{person}{Stelios Andreadis}, \bibinfo{person}{Ilias
  Gialampoukidis}, \bibinfo{person}{Anastasios Karakostas},
  \bibinfo{person}{Stefanos Vrochidis}, \bibinfo{person}{Ioannis Kompatsiaris},
  \bibinfo{person}{Roberto Fiorin}, \bibinfo{person}{Daniele Norbiato}, {and}
  \bibinfo{person}{Michele Ferri}.} \bibinfo{year}{2020}\natexlab{}.
\newblock \showarticletitle{The flood-related multimedia task at mediaeval
  2020}. In \bibinfo{booktitle}{{\em Proceedings of the MediaEval 2020
  Workshop, Online}}. \bibinfo{pages}{14--15}.
\newblock


\bibitem[\protect\citeauthoryear{Bischke, Helber, Schulze, Srinivasan, Dengel,
  and Borth}{Bischke et~al\mbox{.}}{2017}]%
        {bischke2017multimedia}
\bibfield{author}{\bibinfo{person}{Benjamin Bischke}, \bibinfo{person}{Patrick
  Helber}, \bibinfo{person}{Christian Schulze}, \bibinfo{person}{Venkat
  Srinivasan}, \bibinfo{person}{Andreas Dengel}, {and} \bibinfo{person}{Damian
  Borth}.} \bibinfo{year}{2017}\natexlab{}.
\newblock \showarticletitle{The Multimedia Satellite Task at MediaEval 2017.}.
  In \bibinfo{booktitle}{{\em MediaEval}}.
\newblock


\bibitem[\protect\citeauthoryear{Gregory, Payne, McColgin, Cramer, and
  DouglasLove}{Gregory et~al\mbox{.}}{2007}]%
        {Gregory2007}
\bibfield{author}{\bibinfo{person}{Michelle Gregory}, \bibinfo{person}{Deborah
  Payne}, \bibinfo{person}{Dave McColgin}, \bibinfo{person}{Nick Cramer}, {and}
  \bibinfo{person}{V DouglasLove}.} \bibinfo{year}{2007}\natexlab{}.
\newblock \showarticletitle{Visual Analysis of Weblog Content}.
\newblock \bibinfo{journal}{{\em INTER-NATIONAL CONFERENCE ON WEB AND SOCIAL
  MEDIA\/}} (\bibinfo{year}{2007}).
\newblock


\bibitem[\protect\citeauthoryear{Hassan, Ahmad, Hicks, Halvorsen, Al-Fuqaha,
  Conci, and Riegler}{Hassan et~al\mbox{.}}{2020}]%
        {Zohaib2020}
\bibfield{author}{\bibinfo{person}{Syed~Zohaib Hassan}, \bibinfo{person}{Kashif
  Ahmad}, \bibinfo{person}{Steven Hicks}, \bibinfo{person}{P{\aa}l Halvorsen},
  \bibinfo{person}{Ala Al-Fuqaha}, \bibinfo{person}{Nicola Conci}, {and}
  \bibinfo{person}{Michael Riegler}.} \bibinfo{year}{2020}\natexlab{}.
\newblock \showarticletitle{Visual sentiment analysis from disaster images in
  social media}.
\newblock \bibinfo{journal}{{\em arXiv preprint arXiv:2009.03051\/}}
  (\bibinfo{year}{2020}).
\newblock


\bibitem[\protect\citeauthoryear{Khan, Ahmad, Gul, Khan, Ahmad, and
  Al-Fuqaha}{Khan et~al\mbox{.}}{2021}]%
        {khan2021explainable}
\bibfield{author}{\bibinfo{person}{Imran Khan}, \bibinfo{person}{Kashif Ahmad},
  \bibinfo{person}{Namra Gul}, \bibinfo{person}{Talhat Khan},
  \bibinfo{person}{Nasir Ahmad}, {and} \bibinfo{person}{Ala Al-Fuqaha}.}
  \bibinfo{year}{2021}\natexlab{}.
\newblock \showarticletitle{Explainable Event Recognition}.
\newblock \bibinfo{journal}{{\em arXiv preprint arXiv:2110.00755\/}}
  (\bibinfo{year}{2021}).
\newblock


\bibitem[\protect\citeauthoryear{Said, Ahmad, Riegler, Pogorelov, Hassan,
  Ahmad, and Conci}{Said et~al\mbox{.}}{2019}]%
        {said2019natural}
\bibfield{author}{\bibinfo{person}{Naina Said}, \bibinfo{person}{Kashif Ahmad},
  \bibinfo{person}{Michael Riegler}, \bibinfo{person}{Konstantin Pogorelov},
  \bibinfo{person}{Laiq Hassan}, \bibinfo{person}{Nasir Ahmad}, {and}
  \bibinfo{person}{Nicola Conci}.} \bibinfo{year}{2019}\natexlab{}.
\newblock \showarticletitle{Natural disasters detection in social media and
  satellite imagery: a survey}.
\newblock \bibinfo{journal}{{\em Multimedia Tools and Applications\/}}
  \bibinfo{volume}{78}, \bibinfo{number}{22} (\bibinfo{year}{2019}),
  \bibinfo{pages}{31267--31302}.
\newblock


\bibitem[\protect\citeauthoryear{Smith and Fiore}{Smith and Fiore}{2001}]%
        {Smith2001}
\bibfield{author}{\bibinfo{person}{Marc~A Smith} {and}
  \bibinfo{person}{Andrew~T Fiore}.} \bibinfo{year}{2001}\natexlab{}.
\newblock \showarticletitle{Visualization componentsfor persistent
  conversations}.
\newblock \bibinfo{journal}{{\em Proceedings of the SIGCHI conferenceon Human
  factors in computing systems\/}} (\bibinfo{year}{2001}),
  \bibinfo{pages}{136--143}.
\newblock


\bibitem[\protect\citeauthoryear{Syed~Zohaib and Al-Fuqaha}{Syed~Zohaib and
  Al-Fuqaha}{2019}]%
        {Zohaib2019}
\bibfield{author}{\bibinfo{person}{Nicola~Conci Syed~Zohaib, Kashif~Ahmad}
  {and} \bibinfo{person}{Ala Al-Fuqaha}.} \bibinfo{year}{2019}\natexlab{}.
\newblock \showarticletitle{Sentiment Analysis from Images of Natural
  Disasters}.
\newblock \bibinfo{journal}{{\em International Conference on Image Analysis and
  Processing\/}} (\bibinfo{year}{2019}).
\newblock


\end{thebibliography}
\balance
\end{document}